%% file: iclr2027_conference.tex
\documentclass{article} 
\usepackage{iclr2027_conference,times}

\input{math_commands.tex}

\usepackage{hyperref}
\usepackage{url}
\usepackage{xspace}
\usepackage{graphicx}
\usepackage{capt-of}
\usepackage{wrapfig}

\usepackage{booktabs}
\usepackage{multirow}
\usepackage{adjustbox}
\usepackage{makecell}
\usepackage[table]{xcolor}
\definecolor{blue4}{RGB}{85, 142, 213}
\definecolor{red4}{RGB}{255, 102, 102}
\newcommand{\lightred}{\rowcolor{red4!19}}
\newcommand{\lightblue}{\rowcolor{blue4!19}}
\definecolor{linkblue}{RGB}{0,102,204}

\usepackage{listings}
\lstdefinestyle{promptstyle}{
    basicstyle=\ttfamily\footnotesize,
    breaklines=true,
    breakatwhitespace=false,
    frame=single,
    framesep=5pt,
    xleftmargin=5pt,
    xrightmargin=5pt,
    backgroundcolor=\color{gray!10},
    rulecolor=\color{gray!40},
    aboveskip=0pt,
    belowskip=0pt,
    keepspaces=true,
    columns=flexible
}
\usepackage{algorithm}
\usepackage{algpseudocode}

\newcommand{\benchmark}{LME-Bench\xspace}
\newcommand{\benchmarkfull}{Long Multi-turn Image Editing Bench\xspace}
\newcommand{\model}{MT-OPSD\xspace}

\title{On-Policy Self-Distillation for Multi-Turn Image Editing}

\author{
 \textbf{Liangbing Zhao}$^{1}$ ~~ \textbf{Le Zhuo}$^{2}$ ~~ \textbf{Mohamed Elhoseiny}$^{1\dagger}$\\
$^1$ KAUST ~~~ 
$^2$ Krea AI\\[3pt]
{\small $^{\dagger}$Corresponding author}
}

\iclrfinalcopy 
\begin{document}

\maketitle
\lhead{Preprint}

\input{sec/0_abs}
\input{sec/1_intro}
\input{sec/2_related}
\input{sec/3_method}
\input{sec/4_exp}
\input{sec/5_conclusion}

\bibliography{iclr2027_conference}
\bibliographystyle{iclr2027_conference}

\newpage

\appendix
\input{sec/6_appendix}

\end{document}

%% file: math_commands.tex
\usepackage{amsmath,amsfonts,bm}

\def\eqref#1{equation~\ref{#1}}

\def\1{\bm{1}}

\DeclareMathAlphabet{\mathsfit}{\encodingdefault}{\sfdefault}{m}{sl}
\SetMathAlphabet{\mathsfit}{bold}{\encodingdefault}{\sfdefault}{bx}{n}



%% file: sec/0_abs.tex
\begin{abstract}
Instruction-based image editing has achieved strong performance in single-turn settings, yet practical editing is often iterative, with each instruction applied to the output of the previous turn.
We find that existing editing models degrade rapidly under recursive editing and attribute this failure to a train--test mismatch in the conditioning distribution: models are trained on clean source images but must repeatedly condition on their own imperfect outputs at inference time.
To address this, we propose \model, an on-policy self-distillation framework that trains the model on self-generated conditioning states with editing supervision from a clean-conditioned teacher, without requiring multi-turn annotations.
We further introduce \benchmark, a benchmark of 100 ten-turn editing sessions for evaluating long-horizon robustness.
Experiments across three editing backbones show that \model substantially improves long-horizon editing success and reduces multi-turn collapse while largely preserving single-turn editing quality.

\textbf{Project page:} {\textcolor{linkblue}{https://liangbingzhao.github.io/MT-OPSD/}}
\end{abstract}

%% file: sec/1_intro.tex

\section{Introduction}
\label{sec:intro}

Recent progress in instruction-based image editing~\citep{brooks2023instructpix2pix, wei2025omniedit, wu2025qwen} has made it possible to modify images through natural-language instructions, covering a broad range of operations from local object edits to global appearance changes.
Despite strong performance on single-turn edits, real-world editing workflows often involve multiple rounds of refinement.
For example, a user may adjust the lighting of a scene, modify its color palette, reposition an object, and apply a stylistic transformation, with each edit operating on the result of the previous one.
A practical image editor should therefore remain reliable across multiple editing turns, maintaining visual coherence while accurately following each new instruction.

In practice, however, current editing models often struggle across consecutive editing turns.
As a model repeatedly edits its own outputs, small errors accumulate and progressively degrade image quality.
After only a few turns, outputs may exhibit high-frequency chromatic noise, structural fragmentation, or severe identity drift.
We observe this behavior across different editing models, suggesting that multi-turn degradation is a broader limitation of the standard single-turn training paradigm rather than a model-specific failure.

\input{figuretex/noop}
We hypothesize that this degradation arises from a train--test mismatch in the conditioning distribution.
During training, editing models are conditioned on clean source images, whereas at inference time they must operate on their previous outputs, which inevitably contain small model-induced errors.
Although these errors may be visually negligible after a single edit, they accumulate as each output becomes the input to the next turn.
Figure~\ref{fig:noop_drift} provides a simple diagnostic: even when instructed to reproduce the input image without modification, the model exhibits measurable drift at each turn, indicating that recursive editing introduces errors that are not captured by single-turn evaluation.
This behavior is analogous to exposure bias~\citep{huang2026self} in autoregressive video generation, where models trained on ground-truth data are evaluated on their own outputs.

One possible solution is to train directly on multi-turn editing sequences, but such data is scarce.
More importantly, optimizing end-to-end through model-generated multi-turn rollouts would require backpropagation across hundreds of denoising steps and multiple editing turns, making it prohibitively expensive.
Training-free methods such as Emu Edit~\citep{sheynin2024emu} provide another option by reverting nearly unchanged pixels to each turn's input, which can reduce drift for local edits.
However, this approach does not extend to global transformations such as restyling or relighting, where most pixels are expected to change.

To address this gap, we propose \model, an on-policy self-distillation framework for robust multi-turn image editing without requiring multi-turn annotations or ground-truth edited images.
\model builds on the observation that a pretrained editor already exhibits strong single-turn editing behavior under clean conditioning; multi-turn editing introduces an additional challenge as model-induced errors are repeatedly carried into subsequent turns.
Rather than modeling the full distribution of multi-turn editing histories, we isolate this error component through identity rollouts, which preserve the intended image content while accumulating errors from the model's own predictions.
Training then alternates between two complementary objectives: an identity branch that prevents further drift, and an editing branch that pairs these self-generated states with real editing instructions and transfers the model's clean-condition editing behavior through on-policy velocity matching.
An adaptive rollout curriculum progressively exposes the student to deeper self-generated states, while gated teacher promotion updates the clean-condition reference as training proceeds.

To facilitate the evaluation of long-horizon editing robustness, we further construct \benchmarkfull (\benchmark), an evaluation benchmark consisting of 100 editing sessions, each containing 10 consecutive turns with a diverse combination of local and global operations.
Each session is evaluated at every turn in terms of editing accuracy, visual consistency, and image quality, enabling a systematic analysis of when and how multi-turn degradation emerges.

In summary, our contributions are as follows:
\begin{itemize}
\item We show that multi-turn collapse occurs across modern image editing models and attribute it to the train--test mismatch in the conditioning distribution.
\item We propose \model, an on-policy self-distillation framework that transfers the model's own clean-conditioned editing behavior to self-generated states, without requiring multi-turn annotations or an external teacher.
\item We introduce \benchmark, a benchmark of 100 ten-turn editing sessions covering both local and global operations for evaluating long-horizon editing robustness.
\item Experiments across three editing backbones and multiple benchmarks show that \model substantially improves long-horizon editing success and reduces multi-turn collapse while largely preserving single-turn editing quality.
\end{itemize}

%% file: figuretex/noop.tex
\begin{wrapfigure}{r}{0.5\textwidth}
    \centering
    \vspace{-5mm}
    \includegraphics[width=\linewidth]{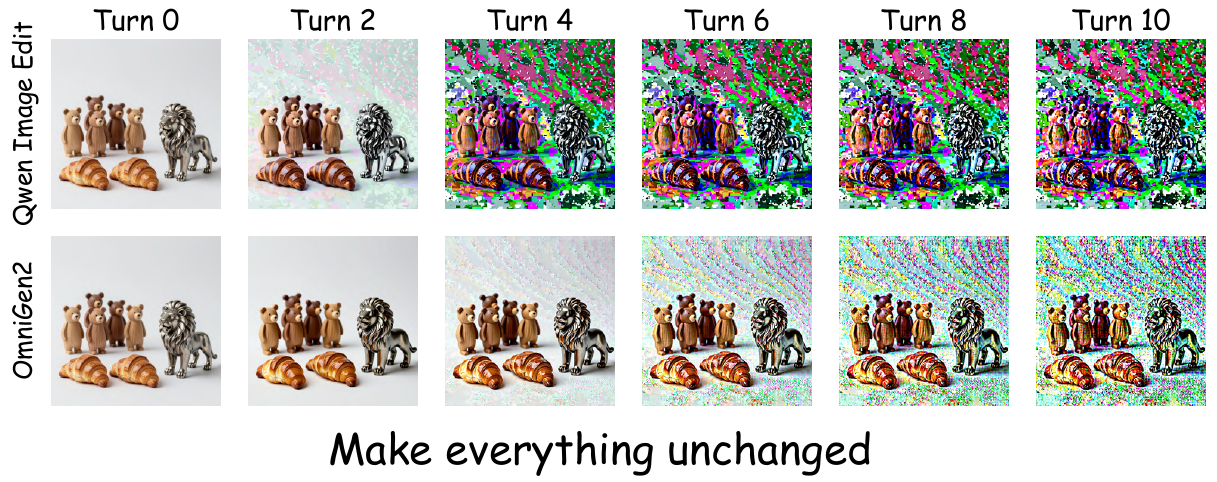}
    \vspace{-5mm}
    \caption{
        \textbf{Drift under identity editing.}
        Repeatedly instructing a model to leave the image unchanged exposes model-induced errors that accumulate across turns, and similar degradation occurs across different editing models.
    }
    \vspace{-5mm}
    \label{fig:noop_drift}
\end{wrapfigure}

%% file: sec/2_related.tex
\section{Related Work}
\label{sec:related}

\paragraph{Instruction-based Image Editing.}
The field of image editing has witnessed a paradigm shift from domain-specific generative adversarial networks~\citep{goodfellow2020generative} to high-fidelity diffusion models~\citep{ho2020denoising, rombach2022high, abdelrahman2025toddlerdiffusion}.
Early diffusion-based approaches~\citep{hertz2022prompt, mokady2023null, zhao2023cross} explored attention manipulation and latent inversion to enable image modifications while preserving the original content, but often struggled with complex and diverse editing instructions.
This motivated instruction-based editing, pioneered by InstructPix2Pix~\citep{brooks2023instructpix2pix} and subsequently advanced through improved data curation and scaling~\citep{zhang2023magicbrush, wei2025omniedit, zhuo2025reflection}, stronger vision-language understanding~\citep{wu2025qwen, wu2025omnigen2, zhao2026statics}, unified generative frameworks~\citep{deng2025emerging, xie2024show}, and in-context flow models~\citep{labs2025flux, liu2025step1x}.
However, these approaches are primarily designed and evaluated for independent single-turn edits, leaving the robustness of models under repeated self-conditioned editing largely unexplored.

\paragraph{Multi-turn Image Editing.}
Recent works have explored improving consistency across sequential edits.
Training-free approaches, including Emu Edit~\citep{sheynin2024emu}, FreqEdit~\citep{liao2026freqedit}, and VAE-LFA~\citep{wang2026dit}, alleviate accumulated degradation through image-space or latent-space corrections, but rely on assumptions about the edits and do not alter the model's editing behavior.
MTC~\citep{zhou2025multi} uses per-turn inversion with trajectory control and adaptive attention guidance on text-to-image models.
VINCIE~\citep{qu2026vincie} and AnchorEdit~\citep{xu2026anchoredit} instead train dedicated models for causal multi-turn editing by adapting video architectures to interleaved image sequences.
Edit-R2~\citep{ye2026edit} focuses on preserving session-level constraints across turns.
Most closely related to our work, MT-EditFlow~\citep{huang2026mt} also attributes multi-turn degradation to exposure bias, but addresses it through reinforcement learning with external reward supervision.
In contrast, \model starts from the observation that pretrained editors already possess strong single-turn editing ability, and uses the model's own clean-condition behavior to extend this ability to self-generated states through on-policy self-distillation.

\paragraph{On-Policy Self-Distillation.}
On-policy self-distillation (OPSD) trains a model on states generated by its own policy while using the same model as teacher under richer conditioning.
In language models, this is commonly realized by providing the teacher with privileged context~\citep{zhao2026self,penaloza2026privileged,sang2026policy}.
In visual generation, D-OPSD~\citep{jiang2026d} conditions the teacher on a paired target image while supervising the student along its own diffusion trajectory, whereas OPSD-V~\citep{liu2026opsd} uses real long-video context to supervise autoregressive generation from self-generated history.
DiffusionOPSD~\citep{zhou2026policy} derives its targets from reward gradients rather than privileged context.
Closely related OPD methods for diffusion and flow models likewise provide teacher supervision along student-generated sampling trajectories~\citep{li2026diffusionopd,fang2026flow,zhou2026danceopd}.
Our setting introduces a different form of on-policy state: in multi-turn editing, the conditioning image itself evolves through the model's previous outputs.
\model therefore uses the clean source image as privileged teacher context and the self-generated rollout state as student context, extending the model's clean-condition editing behavior to recursive editing without paired target images or multi-turn annotations.

%% file: sec/3_method.tex
\section{Method}
\label{sec:method}

\subsection{Preliminaries and Problem Formulation}
\label{sec:formulation}

\paragraph{Multi-turn Image Editing.}
Modern instruction-based editing models are commonly built on flow matching~\citep{lipman2022flow}, where a velocity model $v_\theta(x_t,t,I,e)$ learns a time-dependent vector field between the target edit and Gaussian noise, conditioned on a source image $I$ and an editing instruction $e$.
Given an initial image $I^{(0)}$ and instructions $e_1,\dots,e_K$, multi-turn editing recursively applies the editing operator $G_\theta(I,e)$:
\begin{equation}
I^{(k)}=G_\theta(I^{(k-1)},e_k).
\label{eq:multiturn}
\end{equation}
This multi-turn process introduces a train--test mismatch in the conditioning distribution.
While training exposes the model only to clean source images, later turns condition on self-generated outputs $I^{(k-1)}$.
Such states contain both the intended semantic changes introduced by earlier edits and model-induced errors accumulated across previous editing turns.
The former are part of the editing task itself, while the latter are absent from clean single-turn training and constitute the additional source of mismatch that we target.

\paragraph{On-Policy Self-Distillation.}
On-policy self-distillation (OPSD) combines on-policy distillation~\citep{agarwal2024policy} with self-distillation: the student is supervised at states generated by its own policy, while the same model serves as the teacher under a more informative context~\citep{zhao2026self,penaloza2026privileged}.
Let $s$ denote a state visited along the student rollout, and let $c_S$ and $c_T$ denote the student and teacher contexts, respectively.
For a student $\pi_{\theta}$ and a teacher $\pi_{\bar{\theta}}$ derived from the same model, OPSD minimizes
\begin{equation}
\mathcal{L}_{\mathrm{OPSD}}
=
\mathbb{E}_{s \sim d_{\pi_\theta}(\cdot \mid c_S)}
\left[
D\!\left(
\pi_{\bar{\theta}}(\cdot \mid s,c_T),
\pi_{\theta}(\cdot \mid s,c_S)
\right)
\right],
\label{eq:opsd}
\end{equation}
where $d_{\pi_\theta}$ denotes the state distribution induced by the student and $D$ is a distillation divergence.
This allows the model to distill information available under a richer context into its behavior under a less informative student context, without a separate teacher.
For diffusion models, OPSD is typically instantiated by constructing asymmetric conditioning contexts for the teacher and student, so that predictions under the teacher context provide supervision along the student's denoising process.

\subsection{\model}
\label{sec:alignment}
\input{figuretex/method}

To address the conditioning mismatch, \model isolates the model-induced error component and uses on-policy self-distillation to preserve the model's clean-condition editing behavior on self-generated states.
We denote the student by $\theta_S$ and the teacher by $\theta_T$, both initialized from the same pretrained model.
The student operates on self-generated states, while the teacher is conditioned on the clean source and remains fixed between gated promotions.
The framework consists of four components: self-generated rollout states, a two-branch training objective, an adaptive rollout curriculum, and gated teacher promotion.
The overall framework and training procedure of \model are summarized in Figure~\ref{fig:method} and Algorithm~\ref{alg:mopd}.

\paragraph{Self-Generated Rollout States.}
To obtain conditioning states that contain errors induced by the model itself, we recursively apply the current student to its own outputs.
Specifically, we define an identity instruction $e_{\mathrm{id}}$ (e.g., ``Make everything unchanged'') that asks the model to reproduce the input image without modification, and roll out the current student for $k$ turns starting from a clean source image $I^{(0)}$:
\begin{equation}
\tilde{I}^{(k)}
=
G_{\theta_S}(\tilde{I}^{(k-1)},e_{\mathrm{id}}),
\qquad
\tilde{I}^{(0)}=I^{(0)}.
\end{equation}
The rollout follows the same sampling configuration as inference, so the resulting drift arises from the model's own generation process.
Since the intended image content remains unchanged, the difference between $\tilde{I}^{(k)}$ and $I^{(0)}$ mainly reflects model-induced errors accumulated across the rollout.
Using actual editing instructions during the rollout would instead entangle these errors with intended semantic changes, making it difficult to obtain a reliable supervision signal for learning robustness to model-induced errors.
The identity rollout therefore provides a controlled way to construct on-policy conditioning states from single-turn training data while isolating the error component.

\paragraph{Two-Branch Training Objective.}
Given a rollout state $\tilde{I}^{(k)}$, we optimize two complementary branches, sampled at a fixed ratio.
The identity branch operates under $e_{\mathrm{id}}$ and suppresses unintended changes and accumulated errors across turns, while the editing branch pairs the same rollout state with a real editing instruction $e$ to preserve editing capability under self-generated conditioning.

The identity branch supervises the model under the identity instruction.
A natural choice is to use the clean source image $I^{(0)}$ as the target, asking the model to recover the clean image from its degraded input.
However, we find this restoration objective difficult to optimize, as it requires correcting errors accumulated over $k$ turns within a single turn.
We therefore use the rollout state itself as the target, yielding an identity objective that prevents further drift rather than restoring the clean source:
\begin{equation}
\mathcal{L}_{\mathrm{id}}
=
\mathbb{E}_{t,\epsilon}
\left[
\left\|
v_{\theta_S}(\tilde{x}_t,t,\tilde{I}^{(k)},e_{\mathrm{id}})
-
(\epsilon-\tilde{x}_0)
\right\|_2^2
\right],
\label{eq:loss_id}
\end{equation}
where $\tilde{x}_0$ denotes the latent representation of $\tilde{I}^{(k)}$, $\tilde{x}_t=(1-t)\tilde{x}_0+t\epsilon$, and $\epsilon\sim\mathcal{N}(0,\mathbf{I})$.
This objective prevents accumulated errors from being further amplified, but provides no supervision for executing non-identity edits on degraded states.

The editing branch addresses this limitation by maintaining the model's editing capability under self-generated conditioning.
Because the identity rollout preserves the intended content of $I^{(0)}$, the rollout state $\tilde{I}^{(k)}$ and the clean source differ primarily in accumulated model-induced errors.
For the same editing instruction $e$, we therefore use the model's clean-conditioned prediction as a reference for editing $\tilde{I}^{(k)}$.
Specifically, the teacher $\theta_T$ is conditioned on $I^{(0)}$, while the student $\theta_S$ is conditioned on $\tilde{I}^{(k)}$.
Following the sparse query-based velocity matching of DanceOPD~\citep{zhou2026danceopd}, the student performs an $N$-step denoising process.
We sample a small number of query steps from $p_q$ and match the teacher and student velocities at the corresponding student states:
\begin{equation}
\mathcal{L}_{\mathrm{edit}}
=
\mathbb{E}_{q\sim p_q}
\left[
\left\|
v_{\theta_S}(\bar{x}_{t_q},t_q,\tilde{I}^{(k)},e)
-
v_{\theta_T}(\bar{x}_{t_q},t_q,I^{(0)},e)
\right\|_2^2
\right],
\label{eq:loss_edit}
\end{equation}
where $\bar{x}_{t_q}=\operatorname{sg}(x_{t_q})$ denotes the stop-gradient state reached by the student at query step $q$.
At each queried step, teacher and student are evaluated at the same noisy latent with the same classifier-free guidance (CFG) scale, and each retains its own image condition in both CFG branches, dropping only the text condition in the unconditional branch.

\paragraph{Rollout Curriculum.}
The difficulty of both training branches increases with rollout depth $k$, as deeper rollouts accumulate larger model-induced errors.
Starting from a large depth therefore exposes the model to heavily degraded states early in training and can destabilize optimization.
We instead increase the rollout depth progressively.
Specifically, we measure rollout drift as the mean pixel deviation between $\tilde{I}^{(k)}$ and $I^{(0)}$, and increase the depth by one turn only when the drift remains below a threshold for a fixed number of consecutive training steps.
After each increase, the resulting rise in drift delays further progression until the model adapts to the current depth, yielding an adaptive curriculum without a manually specified schedule.
We cap the rollout depth at $K_{\max}$.
In practice, the adaptive curriculum typically saturates at around four turns, well below this limit.

\paragraph{Gated Teacher Promotion.}
\label{sec:promotion}
The editing branch relies on $\theta_T$ as a stable clean-condition reference.
The pretrained initialization provides a strong starting teacher because it already exhibits reliable single-turn editing behavior under clean conditioning.
As the student is trained on self-generated states with the two-branch objective, however, it can gradually acquire greater robustness to self-generated conditioning than the current teacher.
Continuing to match the same fixed teacher can then limit further improvement.
We therefore allow improved student checkpoints to replace the teacher, while keeping the teacher fixed between promotions.
Each candidate is evaluated asynchronously by a VLM judge on a held-out gate set and promoted only when it satisfies a long-horizon criterion based on success and collapse rates.
The VLM judge is used only for checkpoint selection rather than as a training target or reward signal.
The full promotion rule is provided in Appendix~\ref{sec:app_promotion}.

%% file: figuretex/method.tex
\begin{figure}[t]
    \centering
    \includegraphics[width=\textwidth]{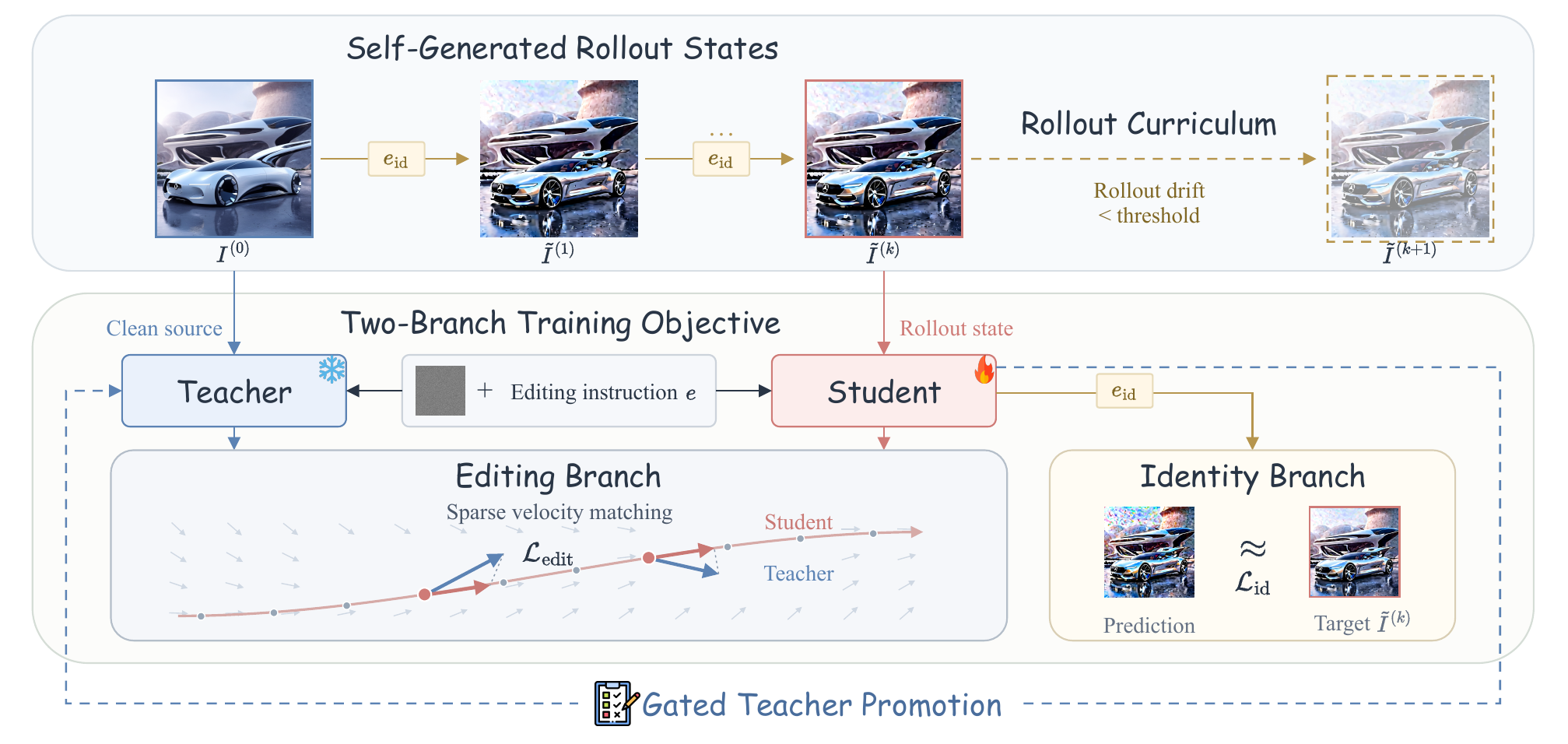}
    \vspace{-3mm}
    \caption{
        \textbf{Overview of \model.}
        The student first constructs self-generated rollout states by repeatedly applying the identity instruction.
        For each rollout state $\tilde{I}^{(k)}$, the identity branch uses the state itself as the target under $\mathcal{L}_{\mathrm{id}}$, while the editing branch aligns the student conditioned on $\tilde{I}^{(k)}$ with a frozen teacher conditioned on the clean source $I^{(0)}$ via sparse query-based velocity matching $\mathcal{L}_{\mathrm{edit}}$.
        A rollout curriculum progressively increases the rollout depth, and improved student checkpoints replace the teacher through gated promotion.
    }
    \label{fig:method}
    \vspace{-3mm}
\end{figure}

%% file: sec/4_exp.tex
\section{Experiments}
\label{sec:exp}

\subsection{\benchmark}
\label{sec:benchmark}

\paragraph{Benchmark Construction.}
Existing multi-turn editing benchmarks contain at most five consecutive turns, making them insufficient to evaluate the long-horizon degradation studied in this work.
We therefore construct \benchmark, which contains 100 editing sessions of 10 consecutive turns each, for a total of 1,000 editing instructions.
Source images are generated by Z-Image-Turbo~\citep{cai2025z} at $1024 \times 1024$ resolution and are evenly distributed across 10 semantic categories.
As shown in Figure~\ref{fig:bench_categories}, each session contains six local edits that modify specific objects or attributes and four global edits that change the overall style, atmosphere, or photometric appearance.
Global edits are placed non-adjacently and are not used in the final turns, so that later local edits are applied to images that have already undergone global changes.
To avoid invalid or ambiguous instructions, we define six families of validity rules, covering cases such as editing invisible regions or requesting a state that is already satisfied.
Candidate instructions are first checked by a VLM and then manually reviewed to ensure that each edit is visible, feasible, and unambiguous.
The full edit taxonomy and composition rules are provided in Appendix~\ref{sec:app_bench_construction}.

\input{figuretex/categories}

\paragraph{Evaluation Protocol and Metrics.}
Each session is executed sequentially, with every turn conditioned only on the output of the previous turn.
We use GPT-4o to evaluate each turn in terms of prompt following, consistency with the previous state, and visual quality.
We report two complementary metrics.
Success rate at turn $k$ (SR@$k$) is the fraction of sessions in which all of the first $k$ turns satisfy both the prompt-following and consistency criteria, reflecting cumulative editing success.
Collapse rate at turn $k$ (CR@$k$) is the fraction of sessions that have undergone persistent visual degradation by turn $k$.
A session is considered collapsed once two consecutive turns are judged visually degraded, and remains counted as collapsed thereafter.
The full judging prompts, thresholds, and evaluation details are provided in Appendix~\ref{sec:app_eval}.

\subsection{Experimental Setup}
\label{sec:setup}

\paragraph{Implementation Details.}
We instantiate \model on three instruction-based editing backbones, including Qwen-Image-Edit-2511~\citep{wu2025qwen}, FLUX.2-klein-base~\citep{labs2025flux2}, and FireRed-Image-Edit~\citep{team2026firered}, and adapt each model using LoRA~\citep{hu2022lora}.
Unless otherwise specified, we use a LoRA rank of 32, set the maximum rollout depth to $K_{\max}=10$, and sample the editing and identity branches at a ratio of 2:1.
Training uses approximately 2,000 source-image/instruction pairs from OmniEdit~\citep{wei2025omniedit}; the corresponding ground-truth edited images are not used.
All self-rollouts follow the same sampling configuration as inference, and GPT-4o is used only for gated teacher promotion during training.
Each run uses four NVIDIA H100/H200 GPUs for training and another four GPUs for asynchronous gate evaluation.
Additional implementation details and hyperparameters are provided in Appendix~\ref{sec:app_detail}.

\paragraph{Evaluation.}
We evaluate long-horizon editing robustness on \benchmark, using SR@$k$ and CR@$k$ as defined in Section~\ref{sec:benchmark}.
We also evaluate on MSE-Bench~\citep{qu2026vincie}, which contains 100 five-turn editing sessions, with each turn applied to the output of the previous one.
Following its official evaluation protocol, GPT-4o scores each turn for prompt following and consistency with the previous state, and a turn is considered successful only when both scores meet the corresponding thresholds.
We report the success rate at each editing turn.
To verify that the improved multi-turn robustness does not sacrifice single-turn editing quality, we further include ImgEdit~\citep{ye2026imgedit}, a standard single-turn instruction-based editing benchmark, and report its overall score.

\paragraph{Comparison Baselines.}
For each backbone, we compare \model against the original base model and three training-free methods for reducing multi-turn degradation.
Emu Edit~\citep{sheynin2024emu} reverts nearly unchanged pixels to each turn's input.
FreqEdit~\citep{liao2026freqedit} mitigates accumulated frequency degradation by recovering high-frequency information, while VAE-LFA~\citep{wang2026dit} corrects low-frequency drift in the VAE latent space.
All methods use the same backbone and inference configuration whenever applicable.
We additionally report GPT-Image-1~\citep{gptimage} and Nano Banana~\citep{nanobanana} as proprietary systems, together with the more recent GPT-Image-2~\citep{gptimage2} and Nano Banana Pro~\citep{nanobananapro} on \benchmark.
Among the training-based multi-turn editing methods discussed in Section~\ref{sec:related}, VINCIE~\citep{qu2026vincie} is the only one with public code and checkpoints.
Since VINCIE is trained as a separate model rather than instantiated on our backbones, we report its comparison separately in Appendix~\ref{sec:app_vincie}.

\subsection{Main Results}
\label{sec:main_results}

\input{tabletex/main_benchmark}
\input{tabletex/main_msebench}

\paragraph{Results on \benchmark.}
Table~\ref{tab:main_bench} shows that all three open-source base models degrade sharply over longer editing sequences, with SR@10 falling to 0.03--0.15 and CR@10 rising to 0.25--0.61.
Existing training-free methods yield model-dependent gains: Emu Edit notably improves Qwen-Image-Edit-2511, VAE-LFA is more effective on FireRed-Image-Edit, while FreqEdit generally degrades long-horizon performance.
In contrast, \model raises SR@10 to 0.38--0.52 across all three backbones and reduces CR@10 to at most 0.04 without sacrificing early-turn performance.
Although the rollout curriculum typically saturates at around four turns, these gains persist through ten-turn evaluation, indicating robustness beyond the rollout depths encountered during training.
Among proprietary systems, GPT-Image-1 also exhibits severe long-horizon degradation and is outperformed by \model across all reported turns and metrics.
Nano Banana rarely collapses, while its lower SR is primarily attributable to failures on global edits that occur predominantly in the earlier turns.
GPT-Image-2 and Nano Banana Pro remain stronger at long horizons.
Overall, these results show that robustness learned from self-generated rollout states transfers to real multi-turn editing sequences, substantially narrowing the long-horizon robustness gap for open-source models.

\paragraph{Results on MSE-Bench and ImgEdit.}
Table~\ref{tab:msebench} reports results on MSE-Bench, an existing five-turn benchmark dominated by local edits, and ImgEdit, which evaluates whether the improved multi-turn robustness preserves single-turn editing quality.
On MSE-Bench, \model improves SR@5 from 0.24 to 0.51 on Qwen-Image-Edit-2511 and from 0.40 to 0.69 on FireRed-Image-Edit, while matching FLUX.2-klein-base at 0.49.
Emu Edit remains competitive in this setting, where the predominance of local edits favors pixel-level restoration, but \model achieves comparable or better performance across the three backbones.
On ImgEdit, \model largely preserves the original single-turn capability, with only minor score changes across all three models.

\paragraph{Qualitative Results.}
Figure~\ref{fig:qualitative} shows selected turns from one ten-turn editing session across three backbones.
The base models produce reasonable results in the early turns, but gradually develop severe visual artifacts as their outputs are repeatedly reused as inputs.
The training-free baselines provide limited improvement: Emu Edit and FreqEdit largely follow the same degradation pattern as the base models, while VAE-LFA better preserves image quality but weakens the requested color-temperature edit at Turn-5.
In contrast, \model remains stable through Turn-10 while following subsequent instructions, showing that long-horizon robustness can be improved without weakening the model's ability to perform later edits.
More detailed results are shown in Appendix~\ref{sec:app_more_qualitative}.

\input{figuretex/qualitative}

\subsection{Ablation Study}
\label{sec:ablation}

\input{tabletex/ablation_components}

All ablations are conducted on Qwen-Image-Edit-2511 under the same training setup.
Table~\ref{tab:ablation} reports the quantitative results, with additional visual comparisons provided in Appendix~\ref{sec:app_ablation_visuals}.

\paragraph{Effect of the Two-Branch Objective.}
The two training branches serve complementary roles.
Removing the editing branch leaves only identity supervision and severely weakens editing ability, with SR@10 falling to 0.00 and CR@10 rising to 0.80.
In contrast, removing the identity branch retains strong long-horizon editing performance and even yields a slightly higher SR@10 of 0.49, but makes the edits less controlled: CR@10 increases from 0.02 to 0.06, SR@1 on MSE-Bench drops from 0.98 to 0.85, and the ImgEdit score falls from 4.49 to 3.88.
Qualitative results in Appendix~\ref{sec:app_ablation_visuals} further show that this variant tends to over-edit, progressively altering untargeted content across turns.
These results suggest that the editing branch preserves editability on self-generated states, while the identity branch suppresses unnecessary changes and preserves untargeted content.

\paragraph{Effect of On-Policy Supervision.}
Replacing on-policy velocity matching with standard flow-matching supervision on teacher-generated pseudo-targets reduces SR@5 from 0.91 to 0.59 and SR@10 from 0.44 to 0.03, while CR@10 rises from 0.02 to 0.67, surpassing the base model.
Standard flow matching provides supervision along interpolation paths toward teacher-generated targets, whereas sparse velocity matching queries the teacher at states visited by the student. 
This suggests that supervision on student-visited states is more effective for long-horizon editing.

\paragraph{Effect of the Rollout Curriculum and Gated Promotion.}
These two components control how the training states and the supervision source evolve during training.
Fixing the rollout depth at four turns reduces SR@10 from 0.44 to 0.17 and increases CR@10 from 0.02 to 0.19, showing that progressive depth expansion is more effective than training at a comparable fixed depth.
Keeping the pretrained initialization as a fixed teacher instead maintains a low CR@10 of 0.06 but limits SR@10 to 0.19.
At the same time, this variant achieves the highest ImgEdit score among the ablations at 4.55, consistent with the view that a fixed clean-condition reference better preserves the pretrained model's single-turn behavior but limits further adaptation to long-horizon editing.

%% file: figuretex/categories.tex
\begin{wrapfigure}{r}{0.5\textwidth}
    \centering
    \vspace{-3mm}
    \includegraphics[width=0.9\linewidth]{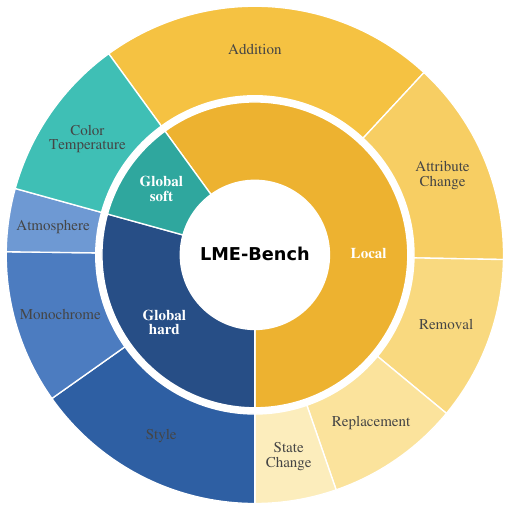}
    \vspace{-2mm}
    \caption{
        Edit-type distribution in \benchmark.
    }
    \label{fig:bench_categories}
    \vspace{-2mm}
\end{wrapfigure}

%% file: tabletex/main_benchmark.tex
\begin{table}[!t]
    \centering
    \caption{
        \textbf{Quantitative comparison on \benchmark.}
        We report the success rate (SR) and the collapse rate (CR) at turns 3, 5, 8, and 10.
    }
    \label{tab:main_bench}
    \renewcommand{\arraystretch}{1.0}
    \setlength{\tabcolsep}{5pt}
    \begin{adjustbox}{max width=\textwidth}
    \begin{tabular}{lcccccccc}
    \toprule
    \multirow{2}{*}{\textbf{Models}} & \multicolumn{4}{c}{\textbf{SR}~$\uparrow$} & \multicolumn{4}{c}{\textbf{CR}~$\downarrow$} \\
    \cmidrule(lr){2-5} \cmidrule(lr){6-9}
     & Turn-3 & Turn-5 & Turn-8 & Turn-10 & Turn-3 & Turn-5 & Turn-8 & Turn-10 \\
    \midrule \lightblue
    \multicolumn{9}{c}{\textbf{\textit{Proprietary Models}}} \\
    \midrule
    GPT-Image-1~\citep{gptimage} & 0.82 & 0.48 & 0.09 & 0.01 & 0.03 & 0.35 & 0.63 & 0.69 \\
    GPT-Image-2~\citep{gptimage2} & 1.00 & 0.98 & 0.98 & 0.96 & 0.00 & 0.00 & 0.00 & 0.00 \\
    Nano Banana~\citep{nanobanana} & 0.67 & 0.66 & 0.63 & 0.56 & 0.00 & 0.00 & 0.00 & 0.00 \\
    Nano Banana Pro~\citep{nanobananapro} & 1.00 & 1.00 & 0.97 & 0.91 & 0.00 & 0.00 & 0.00 & 0.00 \\
    \midrule \lightred
    \multicolumn{9}{c}{\textbf{\textit{Open-Source Models}}} \\
    \midrule
    Qwen-Image-Edit-2511~\citep{wu2025qwen} & 0.98 & 0.43 & 0.08 & 0.03 & 0.00 & 0.10 & 0.45 & 0.55 \\
    + Emu Edit~\citep{sheynin2024emu} & 0.95 & 0.54 & 0.19 & 0.13 & 0.00 & 0.04 & 0.21 & 0.30 \\
    + FreqEdit~\citep{liao2026freqedit} & 0.89 & 0.22 & 0.03 & 0.01 & 0.00 & 0.08 & 0.41 & 0.54 \\
    + VAE-LFA~\citep{wang2026dit} & 0.84 & 0.46 & 0.18 & 0.10 & 0.00 & 0.05 & 0.29 & 0.36 \\
    + \model & \textbf{1.00} & \textbf{0.91} & \textbf{0.63} & \textbf{0.44} & \textbf{0.00} & \textbf{0.01} & \textbf{0.02} & \textbf{0.02} \\
    \midrule
    FireRed-Image-Edit~\citep{team2026firered} & 0.95 & 0.57 & 0.28 & 0.15 & 0.01 & 0.16 & 0.51 & 0.61 \\
    + Emu Edit~\citep{sheynin2024emu} & 0.97 & 0.58 & 0.30 & 0.14 & 0.01 & 0.05 & 0.36 & 0.51 \\
    + FreqEdit~\citep{liao2026freqedit} & 0.93 & 0.41 & 0.04 & 0.00 & 0.01 & 0.11 & 0.64 & 0.73 \\
    + VAE-LFA~\citep{wang2026dit} & 0.88 & 0.70 & 0.50 & 0.37 & 0.01 & 0.08 & 0.26 & 0.33 \\
    + \model & \textbf{0.98} & \textbf{0.88} & \textbf{0.64} & \textbf{0.52} & \textbf{0.00} & \textbf{0.00} & \textbf{0.01} & \textbf{0.03} \\
    \midrule
    FLUX.2-klein-base~\citep{labs2025flux2} & 0.92 & 0.52 & 0.29 & 0.12 & 0.00 & 0.01 & 0.17 & 0.25 \\
    + Emu Edit~\citep{sheynin2024emu} & 0.90 & 0.52 & 0.23 & 0.11 & 0.00 & 0.02 & 0.12 & 0.18 \\
    + FreqEdit~\citep{liao2026freqedit} & 0.64 & 0.19 & 0.03 & 0.01 & 0.00 & 0.03 & 0.35 & 0.54 \\
    + VAE-LFA~\citep{wang2026dit} & 0.81 & 0.51 & 0.31 & 0.16 & 0.00 & 0.03 & 0.16 & 0.24 \\
    + \model & \textbf{0.95} & \textbf{0.76} & \textbf{0.57} & \textbf{0.38} & \textbf{0.00} & \textbf{0.00} & \textbf{0.02} & \textbf{0.04} \\
    \bottomrule
    \end{tabular}
    \end{adjustbox}
    \vspace{-3mm}
\end{table}

%% file: tabletex/main_msebench.tex
\begin{table}[!t]
    \centering
    \caption{
        \textbf{Quantitative comparison on MSE-Bench and ImgEdit.}
        We report the success rate (SR) at each turn on MSE-Bench under its official evaluation protocol, together with the overall score on the single-turn ImgEdit benchmark.
    }
    \label{tab:msebench}
    \renewcommand{\arraystretch}{1.0}
    \setlength{\tabcolsep}{7pt}
    \begin{adjustbox}{max width=\textwidth}
    \begin{tabular}{lcccccc}
    \toprule
    \multirow{2}{*}{\textbf{Models}} & \multicolumn{5}{c}{\textbf{MSE-Bench} (SR~$\uparrow$)} & \multirow{2}{*}{\makecell{\textbf{ImgEdit}\\\textbf{Overall}~$\uparrow$}} \\
    \cmidrule(lr){2-6}
     & Turn-1 & Turn-2 & Turn-3 & Turn-4 & Turn-5 & \\
    \midrule \lightblue
    \multicolumn{7}{c}{\textbf{\textit{Proprietary Models}}} \\
    \midrule
    GPT-Image-1~\citep{gptimage} & 0.96 & 0.69 & 0.67 & 0.64 & 0.56 & 4.20 \\
    Nano Banana~\citep{nanobanana} & 0.99 & 0.77 & 0.75 & 0.73 & 0.63 & 4.29 \\
    \midrule \lightred
    \multicolumn{7}{c}{\textbf{\textit{Open-Source Models}}} \\
    \midrule
    Qwen-Image-Edit-2511~\citep{wu2025qwen} & 0.97 & 0.91 & 0.72 & 0.52 & 0.24 & 4.51 \\
    + Emu Edit~\citep{sheynin2024emu} & \textbf{0.99} & \textbf{0.93} & \textbf{0.80} & 0.69 & 0.49 & 4.36 \\
    + FreqEdit~\citep{liao2026freqedit} & 0.95 & 0.80 & 0.60 & 0.39 & 0.14 & 4.48 \\
    + VAE-LFA~\citep{wang2026dit} & 0.99 & 0.93 & 0.78 & 0.61 & 0.34 & 4.32 \\
    + \model & 0.98 & 0.90 & 0.79 & \textbf{0.69} & \textbf{0.51} & 4.49 \\
    \midrule
    FireRed-Image-Edit~\citep{team2026firered} & \textbf{1.00} & \textbf{0.97} & 0.80 & 0.62 & 0.40 & 4.56 \\
    + Emu Edit~\citep{sheynin2024emu} & 1.00 & 0.95 & 0.86 & 0.77 & 0.59 & 4.33 \\
    + FreqEdit~\citep{liao2026freqedit} & 0.99 & 0.91 & 0.70 & 0.48 & 0.24 & 4.54 \\
    + VAE-LFA~\citep{wang2026dit} & 0.99 & 0.95 & 0.84 & 0.67 & 0.48 & 4.29 \\
    + \model & 0.99 & 0.95 & \textbf{0.87} & \textbf{0.78} & \textbf{0.69} & 4.52 \\
    \midrule
    FLUX.2-klein-base~\citep{labs2025flux2} & \textbf{0.97} & 0.85 & 0.77 & 0.66 & 0.49 & 4.20 \\
    + Emu Edit~\citep{sheynin2024emu} & 0.93 & 0.85 & 0.73 & 0.58 & 0.42 & 4.01 \\
    + FreqEdit~\citep{liao2026freqedit} & 0.85 & 0.57 & 0.35 & 0.19 & 0.11 & 3.91 \\
    + VAE-LFA~\citep{wang2026dit} & 0.95 & 0.86 & 0.69 & 0.52 & 0.41 & 4.08 \\
    + \model & 0.95 & \textbf{0.89} & \textbf{0.77} & \textbf{0.66} & \textbf{0.49} & 4.28 \\
    \bottomrule
    \end{tabular}
    \end{adjustbox}
\end{table}

%% file: figuretex/qualitative.tex
\begin{figure}[ht]
    \centering
    \includegraphics[width=\textwidth]{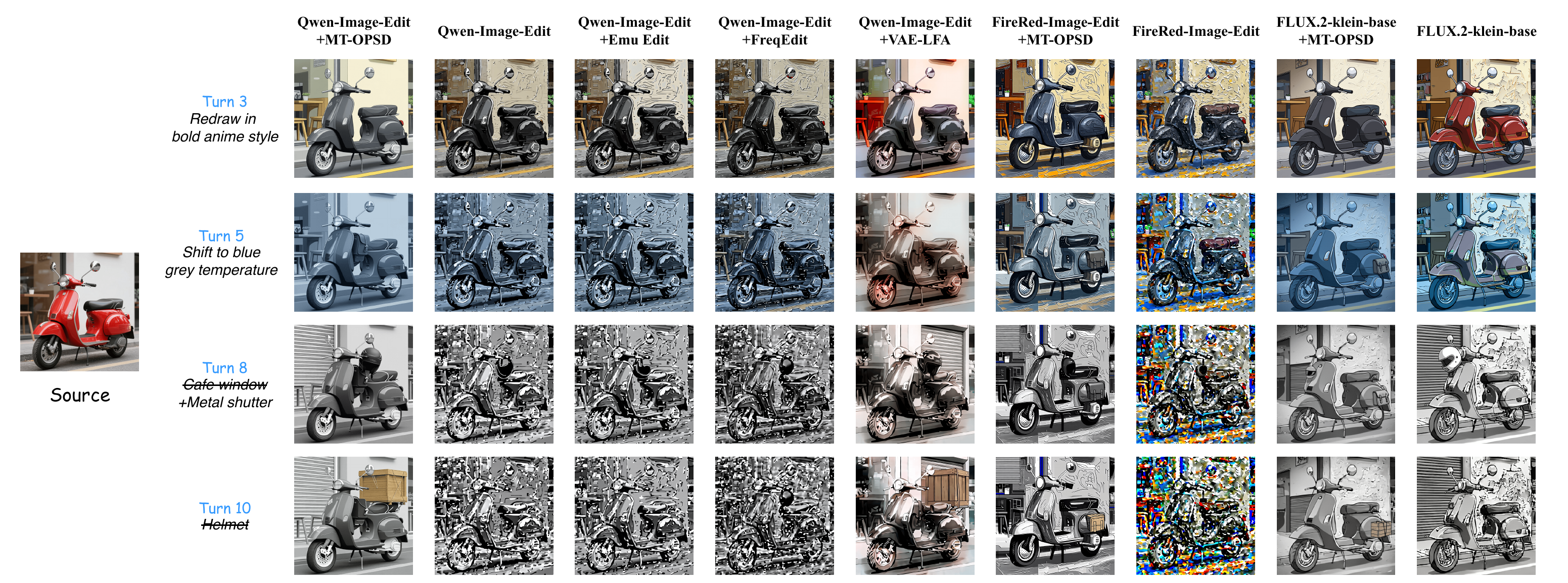}
    \vspace{-2mm}
    \caption{
        \textbf{Qualitative comparison on \benchmark.}
        Repeated editing leads to distinct failure patterns across backbones, while the training-free baselines either inherit these artifacts or weaken the requested edits.
        \model remains stable through 10 turns and continues to follow the editing sequence across all three backbones.
    }
    \label{fig:qualitative}
    \vspace{-3mm}
\end{figure}

%% file: tabletex/ablation_components.tex
\begin{table}[!t]
    \centering
    \caption{
        \textbf{Ablation on key components of \model.} SFT editing supervision replaces on-policy velocity matching with standard flow-matching supervision toward teacher-generated images.
        The curriculum ablation fixes the rollout depth at four, where the adaptive curriculum typically saturates.
    }
    \label{tab:ablation}
    \renewcommand{\arraystretch}{1.0}
    \setlength{\tabcolsep}{6pt}
    \begin{adjustbox}{max width=\textwidth}
    \begin{tabular}{lcccccccc}
    \toprule
    \multirow{2}{*}{\textbf{Method}} & \multicolumn{5}{c}{\benchmark} & \multicolumn{2}{c}{MSE-Bench} & \multirow{2}{*}{\makecell{ImgEdit\\Overall}} \\
    \cmidrule(lr){2-6} \cmidrule(lr){7-8}
     & SR@1 & SR@5 & SR@10 & CR@5 & CR@10 & SR@1 & SR@5 & \\
    \midrule
    Base model & 1.00 & 0.43 & 0.03 & 0.10 & 0.55 & 0.97 & 0.24 & 4.51 \\
    \midrule
    \model & \textbf{1.00} & \textbf{0.91} & 0.44 & \textbf{0.01} & \textbf{0.02} & 0.98 & \textbf{0.51} & 4.49 \\
    w/o identity branch & 0.99 & 0.78 & \textbf{0.49} & 0.03 & 0.06 & 0.85 & 0.50 & 3.88 \\
    w/o editing branch & 0.91 & 0.18 & 0.00 & 0.17 & 0.80 & 0.95 & 0.22 & 4.33 \\
    w/ SFT editing supervision & 1.00 & 0.59 & 0.03 & 0.12 & 0.67 & 0.98 & 0.37 & 4.50 \\
    w/o rollout curriculum & 1.00 & 0.79 & 0.17 & 0.04 & 0.19 & 0.96 & 0.41 & 4.33 \\
    w/o gated promotion & 1.00 & 0.79 & 0.19 & 0.01 & 0.06 & \textbf{1.00} & 0.45 & \textbf{4.55} \\
    \bottomrule
    \end{tabular}
    \end{adjustbox}
\end{table}

%% file: sec/5_conclusion.tex
\section{Conclusion}
\label{sec:conclusion}

In this work, we study the failure of instruction-based editing models under multi-turn editing, where errors in self-generated outputs accumulate into severe degradation.
We attribute this failure to a train--test mismatch in the conditioning distribution and introduce \model, an on-policy self-distillation framework that trains the model on self-generated conditioning states with editing supervision from a clean-conditioned teacher.
We further introduce \benchmark, a benchmark of 100 ten-turn editing sessions for evaluating long-horizon robustness.
Experiments across three editing backbones show that \model improves long-horizon robustness on \benchmark and MSE-Bench while largely preserving single-turn editing quality on ImgEdit.
More broadly, our results suggest that strong single-turn editing capability can be extended to multi-turn settings by learning to operate on self-generated states, without multi-turn annotations.

%% file: sec/6_appendix.tex
\section{Implementation Details}
\label{sec:app_detail}

Table~\ref{tab:app_hparams} lists the full configuration for the three backbones.
LoRA is applied to the attention and feed-forward projections of the diffusion transformer, and all other components remain frozen.
Entries that need further definition are described below.

\begin{itemize}
    \item \textbf{Query distribution.}
    The editing branch uses a training-time trajectory of $N$ steps, independent of the sampling steps used for identity rollout.
    We sample two distinct query positions per update from a Beta distribution over the normalized trajectory index, following DanceOPD~\citep{zhou2026danceopd}.
    Qwen-Image-Edit-2511 and FireRed-Image-Edit use $\mathrm{Beta}(5,5)$, while FLUX.2-klein-base uses $\mathrm{Beta}(2,5)$.
    Unlike DanceOPD, which reports $\mathrm{Beta}(5,2)$ as the best setting, we find queries around the middle or the earlier, noisier part of the trajectory to work better.
    \item \textbf{Identity loss weight.}
    The identity loss is scaled by this factor relative to the editing loss, on top of the branch sampling ratio.
    \item \textbf{Drift threshold and patience.}
    Rollout drift is the mean absolute pixel difference between the current rollout state and the clean source image, computed in the 0 to 255 range.
    Only steps at the current rollout depth contribute.
    The depth increases by one turn once the drift stays at or below the threshold for \emph{patience} consecutive such steps, after which the counter resets.
\end{itemize}

\input{tabletex/appendix/app_hparams}
\section{Training Recipe}
\label{sec:app_algorithm}

Algorithm~\ref{alg:mopd} summarizes the training procedure of \model.
Each step draws a single source image and instruction, constructs the on-policy rollout state with the current student, applies one of the two branches of the training objective, and updates the rollout curriculum.
Teacher promotion runs asynchronously and is described in Appendix~\ref{sec:app_promotion}.

\begin{algorithm}[t]
\caption{Training Procedure of \model.}
\label{alg:mopd}
\begin{algorithmic}[1]
\Require Training set $\mathcal{D}$; pretrained editing model $\theta_0$;
identity instruction $e_{\mathrm{id}}$
\Require Editing probability $\rho$; identity weight $\lambda$;
query distribution $p_q$; initial and maximum rollout depth $k_0$, $K_{\max}$
\Require Drift threshold $\tau$ and patience $P$
\State Initialize student $\theta_S \leftarrow \theta_0$ and teacher $\theta_T \leftarrow \theta_0$
\State Initialize rollout depth $k \leftarrow k_0$ and counter $c \leftarrow 0$

\While{not converged}
    \State Sample $(I^{(0)},e)\sim\mathcal{D}$

    \Statex \textbf{Stage 1: Self-Generated Rollout}
    \State $\tilde I^{(0)} \leftarrow I^{(0)}$
    \For{$j=1,\ldots,k$}
        \State $\tilde I^{(j)}
        \leftarrow
        G_{\theta_S}(\tilde I^{(j-1)},e_{\mathrm{id}})$
        \Comment{no gradient}
    \EndFor

    \Statex \textbf{Stage 2: Two-Branch Training}
    \State Sample $b\sim\mathrm{Bernoulli}(\rho)$
    \If{$b=1$} \Comment{editing branch}
        \State $x_{t_N}\sim\mathcal{N}(0,\mathbf I)$
        \State $\{x_{t_i}\}_{i=0}^{N}
        \leftarrow
        \mathrm{Rollout}(v_{\theta_S};x_{t_N},\tilde I^{(k)},e)$
        \Comment{no gradient}
        \State $\mathcal Q\leftarrow\mathrm{Sample}(p_q,2)$
        \State $\displaystyle
        \mathcal L_{\mathrm{edit}}
        \leftarrow
        \frac{1}{|\mathcal Q|}
        \sum_{q\in\mathcal Q}
        \left\|
        v_{\theta_S}(\bar x_{t_q},t_q,\tilde I^{(k)},e)
        -
        v_{\theta_T}(\bar x_{t_q},t_q,I^{(0)},e)
        \right\|_2^2,$
        \Statex \hspace{\algorithmicindent}
        where $\bar x_{t_q}=\operatorname{sg}(x_{t_q})$
        \State $\mathcal L\leftarrow\mathcal L_{\mathrm{edit}}$
    \Else \Comment{identity branch}
        \State $\tilde x_0\leftarrow\mathrm{Latent}(\tilde I^{(k)})$
        \State Sample $t$ and $\epsilon\sim\mathcal N(0,\mathbf I)$
        \State $\tilde x_t\leftarrow(1-t)\tilde x_0+t\epsilon$
        \State $\displaystyle
        \mathcal L_{\mathrm{id}}
        \leftarrow
        \left\|
        v_{\theta_S}(\tilde x_t,t,\tilde I^{(k)},e_{\mathrm{id}})
        -
        (\epsilon-\tilde x_0)
        \right\|_2^2$
        \State $\mathcal L\leftarrow\lambda\mathcal L_{\mathrm{id}}$
    \EndIf
    \State Update $\theta_S$ by minimizing $\mathcal L$

    \Statex \textbf{Stage 3: Adaptive Rollout Curriculum}
    \State $d\leftarrow
    \operatorname{MeanAbsDiff}(\tilde I^{(k)},I^{(0)})$
    \If{$d\le\tau$}
        \State $c\leftarrow c+1$
    \Else
        \State $c\leftarrow0$
    \EndIf
    \If{$c\ge P$ and $k<K_{\max}$}
        \State $k\leftarrow k+1$; $c\leftarrow0$
    \EndIf

    \Statex \textbf{Stage 4: Gated Teacher Promotion}
    \If{the asynchronous evaluator promotes checkpoint $\theta^\star$}
        \State $\theta_T\leftarrow\theta^\star$
        \Comment{Appendix~\ref{sec:app_promotion}}
    \EndIf
\EndWhile
\end{algorithmic}
\end{algorithm}

\section{Teacher Promotion Rule}
\label{sec:app_promotion}

\paragraph{Gate Set.}
Teacher promotion is determined using a held-out gate set of 23 ten-turn editing sessions.
The sessions are constructed following the same procedure and source-image generator as \benchmark, but use a disjoint set of source images.
Fifteen sessions follow the standard \benchmark composition of six local and four global edits, while the remaining eight contain local edits only.
The gate set is used solely for teacher selection and is excluded from all reported evaluation results.
Checkpoints are evaluated using the same protocol and metrics described in Appendix~\ref{sec:app_eval}.

\paragraph{Asynchronous Evaluation.}
We save a student checkpoint every 50 training samples and evaluate each checkpoint asynchronously on four separate GPUs.
Training proceeds without waiting for evaluation.
Once a checkpoint is promoted, the decision is written to disk and picked up by the training process at its next polling step, where the current teacher is replaced by the promoted checkpoint.

\paragraph{Promotion Rule.}
Each candidate checkpoint is compared with the current teacher at turns 10, 8, and 5, in this order, giving priority to longer editing horizons.
At turn $k$, we define $\Delta\mathrm{SR}_k$ as the number of gate sessions successfully completed through turn $k$ by the candidate minus that of the teacher.
Similarly, $\Delta\mathrm{CR}_k$ is defined as the number of sessions collapsed by turn $k$ under the teacher minus that under the candidate.
Thus, positive values favor the candidate for both quantities.

If $-1 \le \Delta\mathrm{SR}_k < \alpha$ and $-1 \le \Delta\mathrm{CR}_k < \beta$, the comparison is inconclusive at turn $k$, since neither quantity regresses by more than one session and neither reaches its promotion threshold, and the comparison proceeds to the next turn.
Otherwise, the candidate is promoted if either
\begin{align*}
\Delta\mathrm{SR}_k \ge \alpha
\quad &\text{and} \quad
\Delta\mathrm{CR}_k \ge -1
&& \text{(success-led)}, \\
\Delta\mathrm{CR}_k \ge \beta
\quad &\text{and} \quad
\Delta\mathrm{SR}_k \ge 1
&& \text{(stability-led)}
\end{align*}
is satisfied, and the current teacher is retained if neither condition holds.
If all three turns are inconclusive, no promotion is made.
We set $(\alpha,\beta)$ to $(2,5)$ for Qwen-Image-Edit-2511, $(2,3)$ for FLUX.2-klein-base, and $(3,5)$ for FireRed-Image-Edit.
In practice, this rule typically results in one or two teacher promotions during each training run.

\section{\benchmark and Evaluation Details}
\label{sec:app_bench}

This section provides additional details on the construction of \benchmark and its evaluation protocol.

\subsection{Benchmark Construction}
\label{sec:app_bench_construction}

\paragraph{Source Images.}
We generate all source images with Z-Image-Turbo at $1024\times1024$ resolution using 9 sampling steps and no classifier-free guidance.
Each image is seeded by its benchmark index to ensure exact reproducibility.
The 100 sessions are evenly distributed across 10 categories: pets, wild animals, people, food, vehicles, interiors, nature, objects, plants, and urban scenes.

\paragraph{Session Structure.}
Each session contains 10 turns, with six local and four global edits.
Local edits add, remove, replace, or modify a specific object or region, while global edits affect the image as a whole.
We further divide global edits into \emph{hard} globals, which change the rendering medium or apply a strong global transformation such as oil painting, anime, or grayscale, and \emph{soft} globals, which adjust properties such as color temperature or brightness.
Each session contains either three hard and one soft global edit, or two of each.
The first turn is always a hard global edit, no two global edits are adjacent, and the final two turns are local.
This arrangement introduces substantial changes early in the sequence while ensuring that later local edits operate on images that have already undergone several global transformations.
Instructions never require access to an earlier image state or explicit comparison with previous turns; all referenced content must be identifiable from the immediately preceding image.
Table~\ref{tab:app_example_session} shows one complete session.

\paragraph{Validity Rules.}
Unlike single-turn editing, instruction validity in a multi-turn session depends on the preceding edits.
For example, an object referenced by a later instruction may have been removed earlier, or the requested state may already hold.
We therefore construct each session under an explicit set of rules that avoids non-informative turns, where either leaving the image unchanged can satisfy the instruction or success cannot be judged reliably.
The rules are grouped into six families, summarized in Table~\ref{tab:app_rule_families}.
Among them, \textsc{Already-True}, \textsc{Referent}, and \textsc{Interference} specifically address dependencies across editing turns.

\input{tabletex/appendix/app_rule_families}
\input{tabletex/appendix/app_example_session}

\subsection{Evaluation Protocol and Metrics}
\label{sec:app_eval}

\paragraph{Execution.}
Each session is executed sequentially using the standard inference configuration of each backbone.
At every turn, the model receives only the output from the preceding turn as its source image; the original image is never reintroduced.

\paragraph{Instruction Judging.}
We use GPT-4o to evaluate instruction execution at each turn.
Given the source image, all instructions up to the current turn, and the corresponding intermediate outputs, the judge assigns integer scores from 0 to 10 for prompt following and consistency.
Prompt following measures whether the requested change is realized relative to the immediately preceding image, rather than whether the current image merely contains the requested content.
Consistency is defined according to the edit type.
For local edits, it measures unintended changes outside the requested modification; for global edits, it measures whether the main subjects, their identities, and the composition remain consistent under the requested transformation.
A turn is considered successful when both scores are at least 7.

\paragraph{Degradation Judging.}
A separate evaluation pass measures visual degradation independently of instruction execution.
Given the source image, the current output, and the instruction history, the judge scores two aspects from 0 to 2: \emph{content loss}, which measures whether the main subjects and scene remain recognizable, and \emph{surface corruption}, which measures visual artifacts not requested by the instructions.
Requested style changes are explicitly excluded from degradation.
A turn is considered degraded when the two scores sum to at least 2.
The complete judging prompts and scoring criteria are provided in Appendix~\ref{sec:app_judge_prompts}.

\paragraph{Metrics.}
SR@$k$ denotes the fraction of sessions for which all turns up to turn $k$ are successful.
Following the early-stopping protocol of MSE-Bench, instruction evaluation stops after the first failed turn in each session.
CR@$k$ denotes the fraction of sessions that have collapsed by turn $k$.
A session is considered collapsed once two consecutive turns are judged degraded and remains collapsed thereafter.
Requiring two consecutive degraded turns reduces sensitivity to occasional judging noise.
Both metrics are normalized by the total number of sessions.

\subsection{Judge Prompts}
\label{sec:app_judge_prompts}

Listing~\ref{lst:judge_instruction} shows the prompt used for instruction judging and Listing~\ref{lst:judge_degradation} the prompt used for degradation judging.
Fields in braces are filled in per turn.

\begin{lstlisting}[caption={Prompt for instruction judging.}, label={lst:judge_instruction}]
Assume you are an expert in evaluating multi-turn image editing. In this task, a user interacts with an image editing system across multiple turns. At the first turn, the user provides a source image and an editing prompt. The system returns the edited image. In each subsequent turn, the user supplies a new prompt, and the system generates a new image based on the output from the previous turn.
Your goal is to evaluate how successfully the editing instruction of the LAST turn (turn-{num_prompts}) has been executed.

You will be given {num_prompts} user editing prompts and {num_images} images: the first image is the original source image, and the next are the edited results from each turn for each prompt.
You should focus more on the last prompt and the last edited image, but you may also consider the previous prompts and images as context.

The {num_prompts} user editing prompts are: {editing_prompt}

Please follow these evaluation rules. For the LAST turn, assess TWO criteria by giving a reason, then assign an integer score from 0 to 10 for each:

1) prompt_following: does the last edited image fulfill the last user's editing prompt? Judge the DELTA -- compare the last edited image against the IMMEDIATELY PRECEDING image (the second-to-last image provided) and verify that the SPECIFIC change the instruction asks for actually happened, instead of merely checking whether the final image happens to contain the described element. Apply this generally to the instruction type:
- Add a quantity ("add a second / another / one more X"): the COUNT of X must INCREASE by that amount versus the previous image (e.g. one X becomes two). An X that was already there does not count -- there must be a newly added one.
- Remove a quantity ("remove one / a X"): the count of X must DECREASE by one versus the previous image -- not "all of the X disappeared", and not "the X is unchanged".
- Change an attribute ("change X to <color/material/style>"): that specific X must actually take on the new attribute (the existing object changes), not merely that some object with that attribute now appears.
- Add or remove a target: the specific named object must actually appear (for add) or disappear (for remove) versus the previous image.
If the last image is essentially unchanged from the previous image where the instruction demanded a change, prompt_following must be low. (0 indicates the required change did not happen at all; 10 indicates it was executed exactly as asked.)

2) consistency: are the parts that should be preserved still consistent with the input reference (the source image at the first turn, or the result image at the previous turn)?
First decide whether the last instruction is a LOCAL edit (adds, removes, or alters a specific object or region) or a GLOBAL edit (intentionally changes the whole image -- lighting, time of day, weather, season, color grade, or art style, e.g. "make it nighttime", "convert to black and white", "make it a sunset scene").
- LOCAL edit: consistency rates the degree of overediting. 0 indicates the scene is needlessly different from the reference; 10 indicates a minimal yet effective edit that leaves everything else intact.
- GLOBAL edit: the whole image is SUPPOSED to change, so do NOT penalize the global change itself. Instead, consistency rates whether the main subjects, their identities, the composition, and the scene structure are preserved THROUGH the transform. 0 indicates the subjects or composition are lost, replaced, or distorted; 10 indicates the same scene and subjects faithfully re-rendered under the requested global change.

3) Return your results in a JSON structure, following this format:
{"reason": "...", "prompt_following": int, "consistency": int}
\end{lstlisting}

\begin{lstlisting}[caption={Prompt for degradation judging.}, label={lst:judge_degradation}]
You are auditing one frame of a multi-turn image-editing sequence for VISUAL DEGRADATION.

You are given the ORIGINAL SOURCE image first, and the CURRENT image last.
These editing instructions have been applied, in order, to get from the source to the current image:
{instructions}

Some of those instructions deliberately change how the picture LOOKS (repaint as pixel art, as watercolour, as an oil painting, convert to black and white, shift the colour temperature, change the season or weather). A picture that is a COMPETENT rendering in the requested style is NOT degraded, however far it is from a photograph. Blocky flat colour is correct for pixel art; visible brushstrokes are correct for an oil painting. Judge craftsmanship, not photographic realism.

Degradation is the opposite: damage no competent artist would produce. It includes heavy noise, rainbow or colour speckle, blur, melting or duplicated structure, extreme oversaturation, lost detail or sharpness, collapse into noise -- and also: UNREQUESTED flat colour patches replacing texture, hard-edged colour confetti or fragments scattered over the picture, a background that has stopped depicting a place and become an abstract pattern, and a subject reduced to a featureless silhouette.

Answer these two questions about the CURRENT image, then give a brief reason.

1. content_loss -- can the things that should be in the picture still be identified?
   0 = every main object and the setting are identifiable
   1 = at least one main object, or the setting, can no longer be made out
   2 = the picture no longer reads as the scene at all

2. surface_corruption -- is the picture's surface damaged in ways the instructions did not ask for?
   0 = none worth noting
   1 = clearly present, but the picture survives it
   2 = dominates the frame; the frame reads as noise or debris rather than a picture

Return JSON only: {"reason": "...", "content_loss": int, "surface_corruption": int}
\end{lstlisting}

\section{Comparison with VINCIE}
\label{sec:app_vincie}

Among the training-based multi-turn editing methods discussed in Section~\ref{sec:related}, VINCIE~\citep{qu2026vincie} is the only one for which we could obtain both public code and checkpoints.
VINCIE learns in-context editing from video and, in its native setting, conditions each turn on the full editing history, including the clean source image.
We therefore evaluate VINCIE-7B both in this native full-history setting and under the previous-turn protocol used throughout our experiments, where each turn receives only the immediately preceding output.
Table~\ref{tab:app_vincie} compares these settings with the three backbones trained with \model, all evaluated with previous-turn conditioning.

\input{tabletex/appendix/app_vincie}

Under the matched previous-turn protocol, VINCIE degrades sharply on \benchmark: its SR@10 drops from 0.47 under full-history conditioning to 0.09, while CR@10 increases from 0.08 to 0.46.
This suggests that access to the full editing history contributes substantially to its long-horizon performance.
Under the same previous-turn protocol, \model achieves higher SR@5 and SR@10 and lower collapse rates across all three backbones.
Even compared with VINCIE in its native full-history setting, \model obtains higher SR@5 and lower CR@5 and CR@10 on all three backbones, while the SR@10 comparison varies across backbones.
On MSE-Bench, \model matches or exceeds VINCIE at SR@5 on all three backbones.
It also achieves higher single-turn ImgEdit scores than VINCIE.

\section{Additional Qualitative Results}
\label{sec:app_more_qualitative}

\input{figuretex/appendix/appendix_morequali1}
\input{figuretex/appendix/appendix_morequali2}

Figures~\ref{fig:app_morequali1} and~\ref{fig:app_morequali2} show two additional \benchmark sessions, each covering all three backbones and all comparison methods.
Across both sessions, the base models and the three training-free baselines accumulate visible artifacts as editing proceeds, including high-frequency color speckle, fragmented texture, and loss of contrast, and both the dominant failure mode and the turn at which it appears differ across backbones.
\model follows the instruction at every turn on all three backbones, and its outputs remain largely free of such artifacts throughout the sequence.

\section{Qualitative Ablation Results}
\label{sec:app_ablation_visuals}

\input{figuretex/appendix/appendix_ablation}

Figure~\ref{fig:app_ablation} compares \model with the variant trained without the identity branch on two \benchmark sessions.
Both models follow the requested edits throughout the sequence, consistent with the relatively high success rate of the variant in Table~\ref{tab:ablation}.
The main difference lies in the preservation of content that is not targeted by the instructions.
In the first session, the chef's appearance begins to drift from Turn-1 and changes progressively across subsequent edits, leading to a substantial identity shift by Turn-10.
In the second session, the aircraft changes position and orientation from the first turn, with further layout drift accumulating over later turns.
Without the identity branch, repeated editing introduces unnecessary changes to both subject identity and scene layout.
In contrast, \model follows the same editing sequence while better preserving subject identity and the overall composition across all ten turns.
These observations are consistent with the lower single-turn performance of the variant on MSE-Bench and ImgEdit reported in Table~\ref{tab:ablation}.

%% file: tabletex/appendix/app_hparams.tex
\begin{table}[h]
    \centering
    \caption{\textbf{Implementation details of \model across the three backbones.}}
    \label{tab:app_hparams}
    \begin{adjustbox}{max width=\textwidth}
    \begin{tabular}{lccc}
    \toprule
     & Qwen-Image-Edit-2511 & FireRed-Image-Edit & FLUX.2-klein-base \\
    \midrule
    \multicolumn{4}{l}{\textit{LoRA and optimization}} \\
    LoRA rank & 32 & 32 & 32 \\
    LoRA $\alpha$ & 64 & 64 & 64 \\
    Optimizer & AdamW & AdamW & AdamW \\
    Learning rate & $3\times10^{-4}$ & $3\times10^{-4}$ & $1\times10^{-4}$ \\
    Learning-rate schedule & constant & constant & constant \\
    Training samples & 2{,}000 & 2{,}000 & 2{,}000 \\
    Training resolution & $512^2$ & $512^2$ & $512^2$ \\
    \midrule
    \multicolumn{4}{l}{\textit{Identity rollout}} \\
    Sampling steps & 30 & 30 & 50 \\
    Guidance scale & 4.0 & 4.0 & 4.0 \\
    \midrule
    \multicolumn{4}{l}{\textit{Editing branch}} \\
    Trajectory steps $N$ & 10 & 10 & 10 \\
    Query steps & 2 & 2 & 2 \\
    Query distribution $p_q$ & $\mathrm{Beta}(5,5)$ & $\mathrm{Beta}(5,5)$ & $\mathrm{Beta}(2,5)$ \\
    Guidance scale & 4.0 & 4.0 & 4.0 \\
    \midrule
    \multicolumn{4}{l}{\textit{Objective and curriculum}} \\
    Identity loss weight & 10 & 10 & 10 \\
    Initial rollout depth & 2 & 2 & 2 \\
    Maximum rollout depth $K_{\max}$ & 10 & 10 & 10 \\
    Drift threshold & 9.0 & 10.0 & 10.0 \\
    Patience & 10 & 10 & 10 \\
    \bottomrule
    \end{tabular}
    \end{adjustbox}
\end{table}

%% file: tabletex/appendix/app_rule_families.tex
\begin{table}[h]
    \centering
    \caption{\textbf{Validity rules used to construct \benchmark.}
    The rules are grouped into six families according to the type of invalid or non-informative turn they prevent.}
    \label{tab:app_rule_families}
    \begin{adjustbox}{max width=\textwidth}
    \begin{tabular}{lll}
    \toprule
    Family & Invalidity condition & Example rule \\
    \midrule
    \textsc{Structure}
        & The chain violates the session schema
        & No two global edits are adjacent \\
    \textsc{Visibility}
        & The requested change is not visually verifiable
        & Counts in add/remove instructions are at most three \\
    \textsc{Already-True}
        & The requested state already holds
        & A pose edit must request the opposite of the current state \\
    \textsc{Referent}
        & The referenced object cannot be resolved
        & After a container is replaced, later turns may not refer to the old one \\
    \textsc{Interference}
        & Another turn would cancel or invalidate the edit
        & A grayscale or sepia edit must be the last global edit \\
    \textsc{Feasibility}
        & The requested change cannot be rendered reliably in the frame
        & An added object requires sufficient visible space at its implied size \\
    \bottomrule
    \end{tabular}
    \end{adjustbox}
\end{table}

%% file: tabletex/appendix/app_example_session.tex
\begin{table}[h]
    \centering
    \caption{\textbf{Example ten-turn session from \benchmark.}}
    \label{tab:app_example_session}
    \begin{adjustbox}{max width=\textwidth}
    \begin{tabular}{clp{0.72\textwidth}}
    \toprule
    Turn & Type & Instruction \\
    \midrule
    1  & hard global & Repaint the whole image as a textured oil painting with visible brushstrokes, keeping the dog and the room. \\
    2  & local       & Change the wall behind the dog to a warm sage-green color. \\
    3  & hard global & Redraw the whole image in a bold anime illustration style with clean lines and cel shading. \\
    4  & local       & Add a rubber ball on the floor beside the dog. \\
    5  & soft global & Shift the whole image to a cool blue-gray color temperature. \\
    6  & local       & Make the dog lie down on the floor instead of sitting. \\
    7  & hard global & Convert the entire image to black and white, full grayscale with no color. \\
    8  & local       & Change the wooden floor into pale marble tiles. \\
    9  & local       & Add a small potted plant in the corner behind the dog. \\
    10 & local       & Remove the rubber ball from the floor. \\
    \bottomrule
    \end{tabular}
    \end{adjustbox}
\end{table}

%% file: tabletex/appendix/app_vincie.tex
\begin{table}[h]
    \centering
    \caption{
        \textbf{Comparison with VINCIE under different conditioning histories.}
        VINCIE is evaluated in its native full-history setting and under the previous-turn protocol used throughout our experiments.
        All \model variants use previous-turn conditioning.
    }
    \label{tab:app_vincie}
    \renewcommand{\arraystretch}{1.0}
    \setlength{\tabcolsep}{6pt}
    \begin{adjustbox}{max width=\textwidth}
    \begin{tabular}{lcccccccc}
    \toprule
    \multirow{2}{*}{\textbf{Method}} & \multicolumn{5}{c}{\benchmark} & \multicolumn{2}{c}{MSE-Bench} & \multirow{2}{*}{\makecell{ImgEdit\\Overall}} \\
    \cmidrule(lr){2-6} \cmidrule(lr){7-8}
     & SR@1 & SR@5 & SR@10 & CR@5 & CR@10 & SR@1 & SR@5 & \\
    \midrule
    VINCIE-7B (full history) & 0.81 & 0.64 & 0.47 & 0.06 & 0.08 & 0.95 & 0.49 & 3.86 \\
    VINCIE-7B (previous turn) & 0.84 & 0.15 & 0.09 & 0.17 & 0.46 & 0.91 & 0.30 & 3.86 \\
    \midrule
    Qwen-Image-Edit-2511 + \model & \textbf{1.00} & \textbf{0.91} & 0.44 & 0.01 & \textbf{0.02} & 0.98 & 0.51 & 4.49 \\
    FireRed-Image-Edit + \model & 1.00 & 0.88 & \textbf{0.52} & \textbf{0.00} & 0.03 & \textbf{0.99} & \textbf{0.69} & \textbf{4.52} \\
    FLUX.2-klein-base + \model & 0.99 & 0.76 & 0.38 & 0.00 & 0.04 & 0.95 & 0.49 & 4.28 \\
    \bottomrule
    \end{tabular}
    \end{adjustbox}
\end{table}

%% file: figuretex/appendix/appendix_morequali1.tex
\begin{figure}[t]
    \centering
    \includegraphics[height=0.92\textheight]{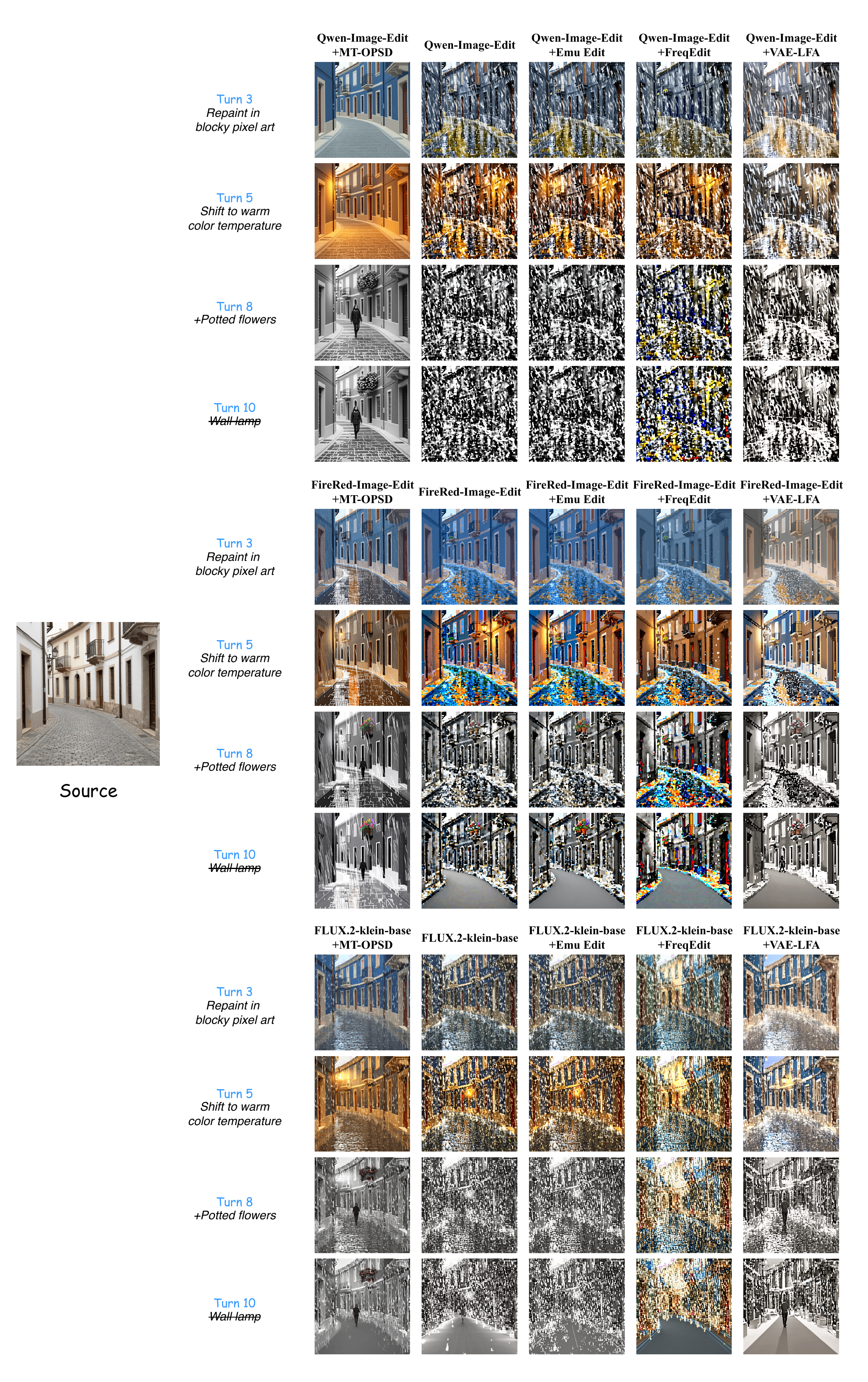}
    \caption{
        \textbf{Additional qualitative comparison on \benchmark.}
        Selected turns of one editing session across three backbones.
    }
    \label{fig:app_morequali1}
\end{figure}

%% file: figuretex/appendix/appendix_morequali2.tex
\begin{figure}[t]
    \centering
    \includegraphics[height=0.92\textheight]{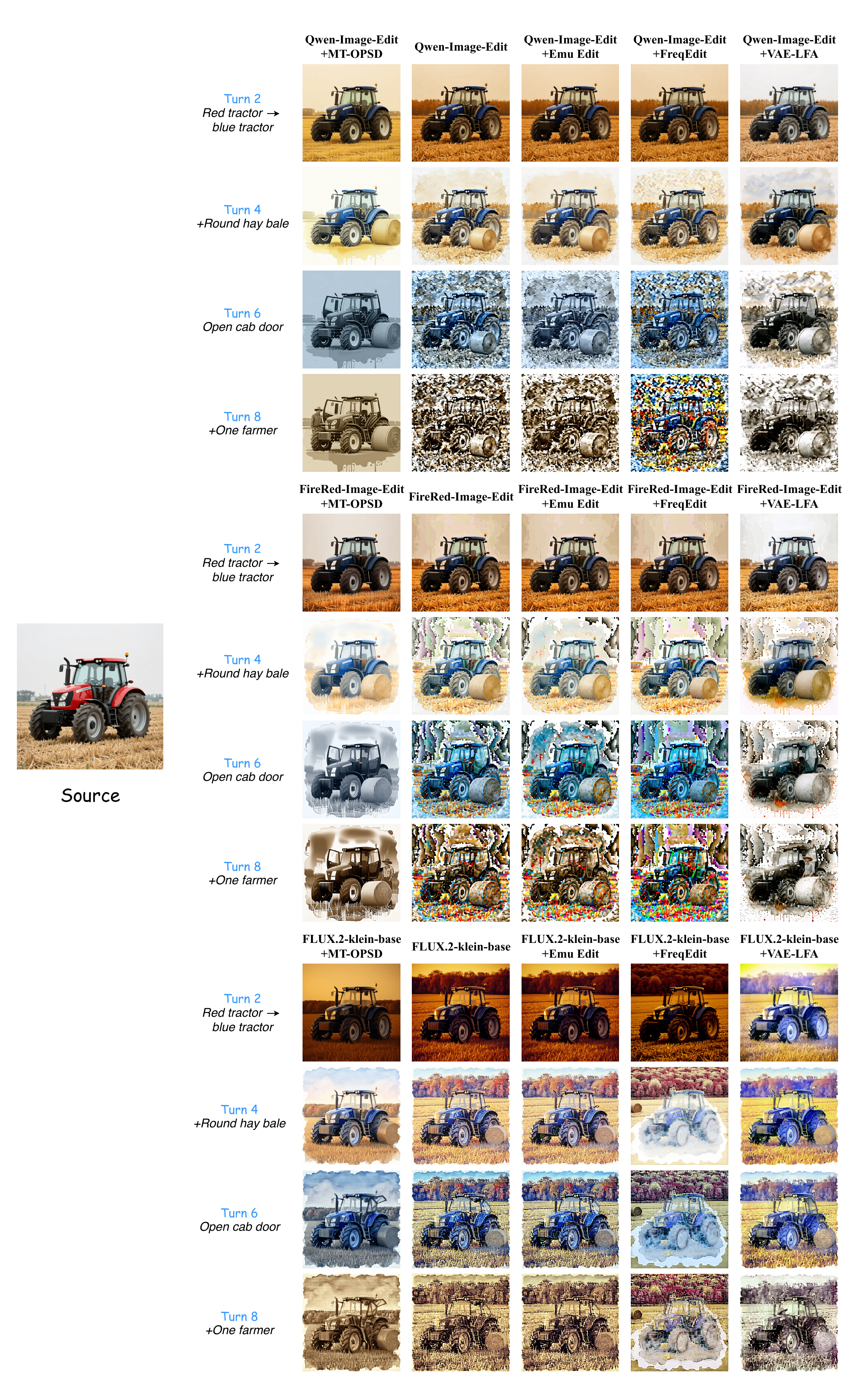}
    \caption{
        \textbf{Additional qualitative comparison on \benchmark.}
        Selected turns of one editing session across three backbones.
    }
    \label{fig:app_morequali2}
\end{figure}

%% file: figuretex/appendix/appendix_ablation.tex
\begin{figure}[t]
    \centering
    \includegraphics[width=\textwidth]{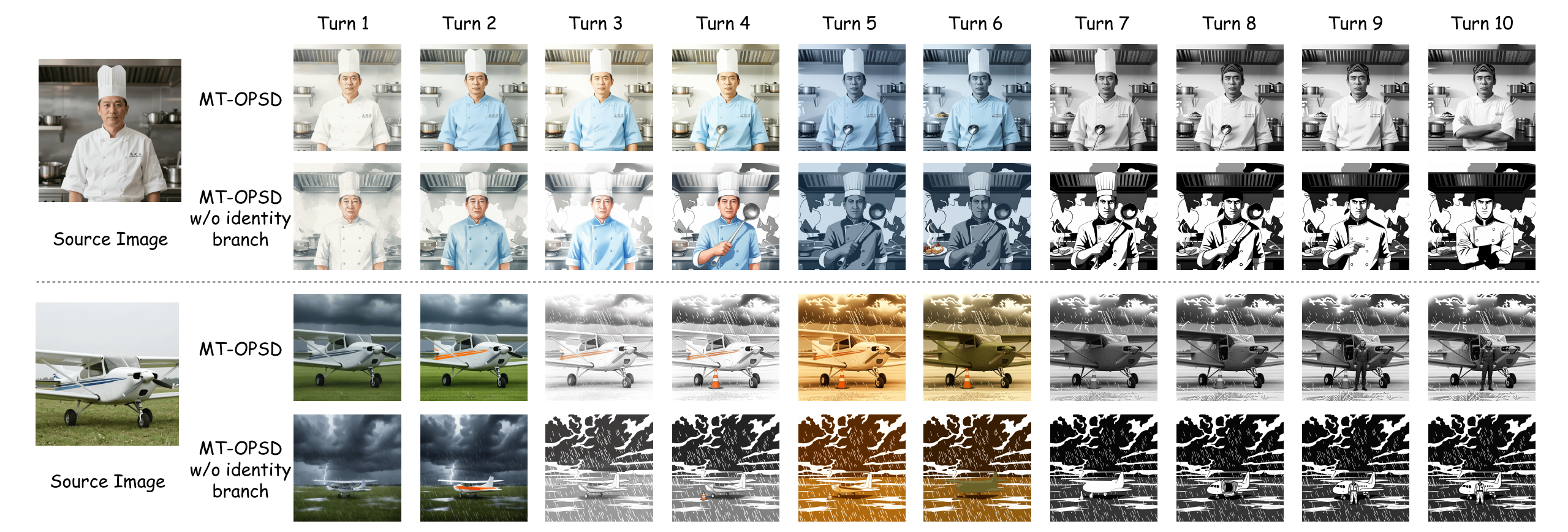}
    \caption{
        \textbf{Effect of the identity branch.}
        Removing the identity branch preserves instruction following but leads to progressive identity and layout drift across turns.
        \model better preserves untargeted content while following the same ten-turn editing sequences.
    }
    \label{fig:app_ablation}
\end{figure}